\documentclass[letterpaper, 10 pt, conference]{ieeeconf}
\IEEEoverridecommandlockouts
\usepackage{multirow}
\usepackage[ruled,vlined]{algorithm2e}
\usepackage{orcidlink}
\usepackage{cite}
\usepackage{amsmath,amssymb,amsfonts}
\usepackage{algorithmic}
\usepackage{graphicx}
\usepackage{textcomp}
\usepackage[shortcuts,acronym]{glossaries}
\usepackage{xcolor}
\usepackage{booktabs}
\usepackage{cleveref}
\usepackage{siunitx}

\usepackage{tikz}

\newcommand{\ieeecopyrightnotice}{%
\begin{tikzpicture}[remember picture,overlay]
\node[
  anchor=south,
  align=center,
  text=black,
  font=\scriptsize,
  inner sep=0pt
] at ([yshift=9mm]current page.south) {%
\begin{minipage}{0.92\textwidth}
\centering
\textcopyright~2026 IEEE. Personal use of this material is permitted.
Permission from IEEE must be obtained for all other uses, in any current
or future media, including reprinting/republishing this material for advertising
or promotional purposes, creating new collective works, for resale or
redistribution to servers or lists, or reuse of any copyrighted component of
this work in other works.
\end{minipage}
};
\end{tikzpicture}%
}

\glsdisablehyper
\newif\ifanonymous
\anonymousfalse %

\ifanonymous
\else
    \hypersetup{
        hidelinks
    }
\fi

\title{\LARGE \bf AerialFusionMapNet: Online HD Map Construction with Aerial-Onboard BEV Fusion
}

\newacronym{hd}{HD}{High-Definition}
\newacronym{hdmap}{HD~map}{High-Definition Map}
\newacronym{bev}{BEV}{Bird’s-Eye View}
\newacronym{lidar}{LiDAR}{Light Detection and Ranging}
\newacronym{uav}{UAV}{Unmanned Aerial Vehicle}
\newacronym{nuscenes}{nuScenes}{nuScenes Autonomous Driving Dataset}
\newacronym{mse}{MSE}{Mean Squared Error}
\newacronym{cnn}{CNN}{Convolutional Neural Network}
\newacronym{nds}{NDS}{nuScenes Detection Score}
\newacronym{gt}{GT}{Ground Truth}
\newacronym{map}{mAP}{mean Average Precision}
\newacronym{roi}{RoI}{Region of Interest}
\newacronym{cka}{CKA}{Centered Kernel Alignment}
\newacronym{sd}{SD}{Standard-Definition}
\newacronym{scr}{SCR}{Scenario-Consistent Rotation}
\newacronym{cvs}{CVS}{Cross-View Supervision}
\newacronym{p2bev}{P2BEV}{Perspective-to-BEV}
\newacronym{ap}{AP}{Average Precision}

\ifanonymous
\else
\author{%
Daniel Lengerer$^{1}$~\orcidlink{0009-0009-1600-6144},
Mathias Pechinger$^{2}$~\orcidlink{0000-0003-2371-9870},
Klaus Bogenberger$^{2}$~\orcidlink{0000-0003-3868-9571},
Carsten Markgraf$^{1}$~\orcidlink{0000-0001-9447-2065}%
\thanks{$^{1}$Technical University of Applied Sciences Augsburg, Germany.
{\tt\small \{Daniel.Lengerer, Carsten.Markgraf\}@tha.de}}%
\thanks{$^{2}$Technical University of Munich, Germany.
{\tt\small \{Klaus.Bogenberger, Mathias.Pechinger\}@tum.de}}%
}
\fi

\begin{document}
\vspace{0.15em}
\maketitle
\ieeecopyrightnotice

\begin{abstract}
High-resolution aerial imagery has recently emerged as a complementary modality for automated driving perception and has shown potential to improve \acrlong{bev} scene understanding when fused with onboard sensors. Prior work demonstrated performance gains for online \acrshort{hd} map construction through aerial-onboard fusion; however, conventional end-to-end fusion does not fully exploit the structural information contained in aerial representations.
In this work, we introduce \emph{AerialFusionMapNet}, a fusion-based mapping framework with a structured two-stage training strategy that explicitly enhances the contribution of aerial features within a unified pipeline. The proposed training scheme enables more effective integration of structural aerial priors.
On the nuScenes geographic split, \emph{AerialFusionMapNet} achieves up to 54.7 \acrshort{map}, improving over prior aerial-onboard fusion baselines (48.8 \acrshort{map}) by +5.9 absolute and +12.1\% relative. The results suggest that structured training design, rather than increased architectural complexity, plays a more decisive role in unlocking the full potential of aerial imagery for online \acrshort{hd} map construction.
Code and trained models are available at \url{https://github.com/DriverlessMobility/AerialFusionMapNet}.
\end{abstract}

\section{Introduction}\label{sec:intro}

\gls{hd} maps provide structured geometric and semantic representations of road environments and are a key component of automated driving systems. They support motion prediction, behavior planning, and scene understanding by encoding lane topology and other static infrastructure elements in a machine-readable form. 

Conventional \gls{hd} maps are generated using dedicated mapping fleets and extensive offline post-processing \cite{hdmapchallenges}, resulting in high accuracy but limited scalability and costly update cycles. 
To address these challenges, online \gls{hd} map construction has been introduced as an alternative approach, in which vectorized map elements are inferred directly from onboard perception during vehicle operation \cite{maptracker,MapTR,Streammapnet,MapTRv2}. 
These methods typically rely on \gls{bev}-based representations derived from multi-camera inputs.

Perception systems that rely predominantly on ego-vehicle sensors can remain constrained by occlusions, limited spatial context, and restricted sensing range, particularly in dense or cluttered urban environments \cite{Streammapnet}. To complement onboard sensing, overhead aerial imagery has recently been explored as an additional structural prior. High-resolution aerial images, acquired offline from satellite or aircraft-based surveys and registered to the vehicle-centric coordinate frame, provide a global view of static road infrastructure. The AID4AD work~\cite{lengerer2025aid4adaerialimagedata} demonstrated that such aerial–onboard fusion can improve \gls{bev}-based online map construction under a fixed fusion setup.
However, the integration of aerial imagery into online mapping remains insufficiently understood. Existing approaches primarily establish feasibility but leave open how aerial representations should be trained, how fusion strategies influence performance, and how robust such systems are to aerial image misalignment.

In this paper, we present \emph{AerialFusionMapNet}, a framework for online \gls{hd} map construction that systematically investigates the integration of aerial imagery into \gls{bev}-based perception. 
Our central observation is that naive end-to-end training does not fully exploit structural information from the aerial view.
We therefore introduce a structured two-stage training strategy that pretrains aerial encoders and applies \gls{cvs}~\cite{cvs_bev_surround_teacher} during joint optimization, enabling more effective cross-view representation alignment.

Extensive experiments on the nuScenes geographic split show that \emph{AerialFusionMapNet} achieves up to 54.7 \gls{map}, clearly outperforming prior aerial–onboard fusion baselines. In addition, we analyze the influence of aerial encoder architecture and evaluate robustness under controlled aerial misalignment, providing practical insights into robust aerial–onboard map construction.

The main contributions of this work are summarized as follows:
\begin{itemize}

    \item We introduce \textbf{AerialFusionMapNet}, a fusion-based framework for online \gls{hd} map construction that integrates offline aerial imagery through a structured two-stage training strategy, enabling effective aerial–onboard feature alignment within a unified \gls{bev} representation.

    \item We conduct a systematic and controlled study of aerial–onboard fusion design, isolating the effects of aerial encoder architecture, aerial-only pretraining, cross-view supervision, and aerial image misalignment, thereby providing principled insights into multimodal map construction beyond prior feasibility-driven setups.

    \item We demonstrate consistent performance gains on the nuScenes geographic split, achieving up to 54.7 \gls{map} and improving over prior aerial–onboard fusion baselines by +12.1\% relative under identical evaluation settings.

\end{itemize}

\begin{figure*}[t]
    \centering
    \includegraphics[width=0.85\textwidth]{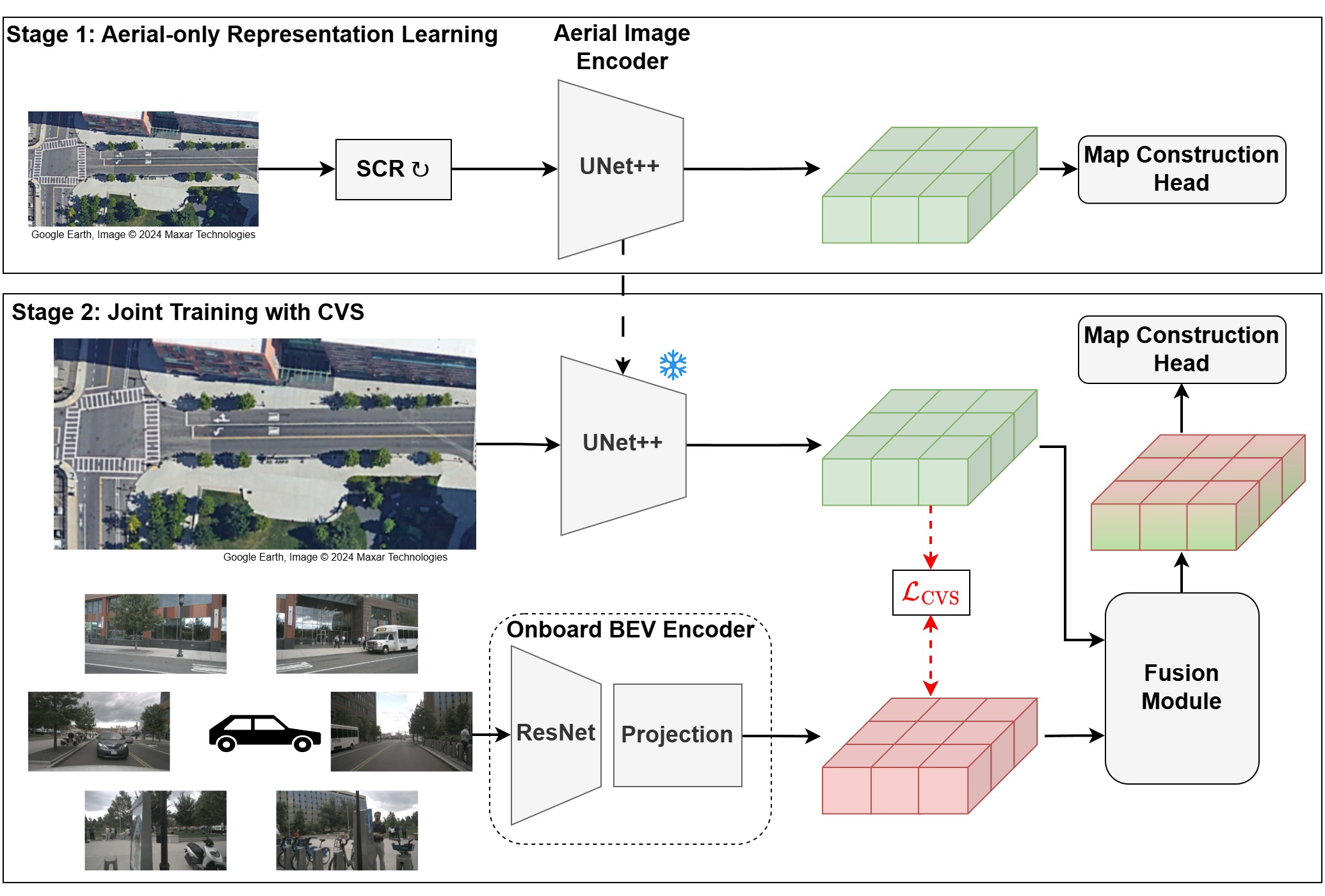}
    \caption{
    Overview of AerialFusionMapNet and its two-stage training strategy.
    Stage~1 performs aerial-only pretraining with \gls{scr}.
    In Stage~2, the pretrained aerial encoder is frozen and aligned with onboard \gls{bev} features via \gls{cvs} before fusion and map decoding.
    }
    \label{fig:afm_overview}
\end{figure*}

\section{Related Work}\label{sec:related}

We organize related work according to the main components of our approach: 
(A) online HD map construction from onboard sensors, 
(B) fusion of offline aerial imagery with vehicle-centric perception, 
(C) auxiliary supervision for perspective-to-BEV representation learning, and 
(D) integration of navigation map priors into \gls{bev} perception.

\subsection{Online HD Map Construction from Onboard Sensors}

Online \gls{hd} map construction has evolved from rasterized \gls{bev} segmentation toward end-to-end vectorized formulations. HDMapNet~\cite{li2021HDMapNet} predicts dense semantic maps requiring post-processing, while VectorMapNet~\cite{liu2022vectormapnet}, MapTR~\cite{MapTR}, and MapTRv2~\cite{MapTRv2} directly generate structured vectorized map elements using query-based decoders. Mask2Map~\cite{Mask2Map} extends these query-based formulations with instance-level mask prediction mechanisms. Streaming and temporally aggregated variants extend these formulations to longer horizons \cite{Streammapnet,zhang2023online_gemap}.

Memory-centric approaches such as MapTracker~\cite{maptracker} further enhance spatial–temporal consistency by modeling map construction as a tracking problem and propagating map elements across frames. Across raster-based, vectorized, streaming, and memory-centric formulations, map construction remains grounded in ego-centric onboard sensing. Our work extends this paradigm by incorporating complementary overhead context into \gls{bev}-based online mapping.

\subsection{Fusion of Offline Aerial Imagery with Vehicle-Centric Perception}

Prior research has investigated aerial or satellite imagery to support lane-level mapping and scene understanding. SatforHDMap~\cite{satforhd} and SIO-Mapper~\cite{cho2025siomapperframeworklanelevelhd} leverage overhead imagery to refine road structure and provide global context, but typically rely on coarse alignment or operate at map level without precise vehicle-centric registration.

AID4AD~\cite{lengerer2025aid4adaerialimagedata} provides accurately registered aerial imagery for nuScenes~\cite{nuScenes}, enabling controlled aerial–onboard fusion experiments. The original study established the potential of incorporating overhead imagery into BEV-based online map construction. In contrast, this work systematically examines how aerial–onboard integration behaves under different encoder choices, training strategies, and alignment conditions.

Several works also extract road topology directly from overhead imagery, including RoadTracer~\cite{roadtracer}, Sat2Graph~\cite{sat2graph}, PolyMapper~\cite{polymapper}, and Pix2Poly~\cite{adimoolam2024pix2poly}. These methods focus on standalone cartographic reconstruction rather than integration with vehicle-centric online perception.

\subsection{Auxiliary Supervision for Perspective-to-BEV Feature Learning}

Learning geometrically consistent \gls{bev} representations from camera inputs remains challenging due to depth ambiguity and projection noise. Distillation-based approaches such as DistillBEV~\cite{Wang2023DistillBEV}, BEV-LGKD~\cite{Li2022BEVLGKD}, and MapDistill~\cite{Hao2024MapDistill} transfer geometric or semantic knowledge from multi-sensor teachers to camera-based encoders. Complementary methods introduce explicit structural guidance; for example, BEVDiffuser~\cite{Ye2025BEVDiffuser} employs diffusion-based, ground-truth layout–conditioned denoising to refine \gls{bev} feature representations.

\Gls{cvs} proposes a representation learning paradigm for feature-level refinement of ego-centric \gls{bev} encoders using geo-aligned overhead imagery as a supervisory signal~\cite{cvs_bev_surround_teacher}. 
One-for-All~\cite{OneForAll} shows that mismatches in feature statistics and inductive biases can hinder direct feature transfer across both homogeneous and heterogeneous teacher–student configurations, motivating lightweight refinement mechanisms. 
In this work, we employ CVS as an auxiliary supervision signal within a fusion-based map construction framework and analyze its role in a staged training procedure.

\subsection{Navigation Map Priors in BEV Perception}

Prior work incorporates navigation-grade \gls{sd} maps as structured priors for \gls{bev} perception. P-MapNet~\cite{pmapnet}, BlosBEV~\cite{blosbev}, RoadPainter~\cite{Ma2023RoadPainter}, and COG-MP~\cite{cogmp} inject rasterized or structured map information into vectorized mapping pipelines, while NavMapFusion~\cite{monninger2025navmapfusion} conditions a diffusion-based decoder on navigation maps.

Navigation priors provide compact, symbolic road layout information. In contrast, dense overhead imagery offers visually grounded structural cues with different fusion characteristics. Our work investigates this complementary modality within a unified online mapping framework.

\begin{figure*}[t]
\centering

\begin{minipage}[t]{0.36\linewidth}
    \centering
    \includegraphics[width=\linewidth]{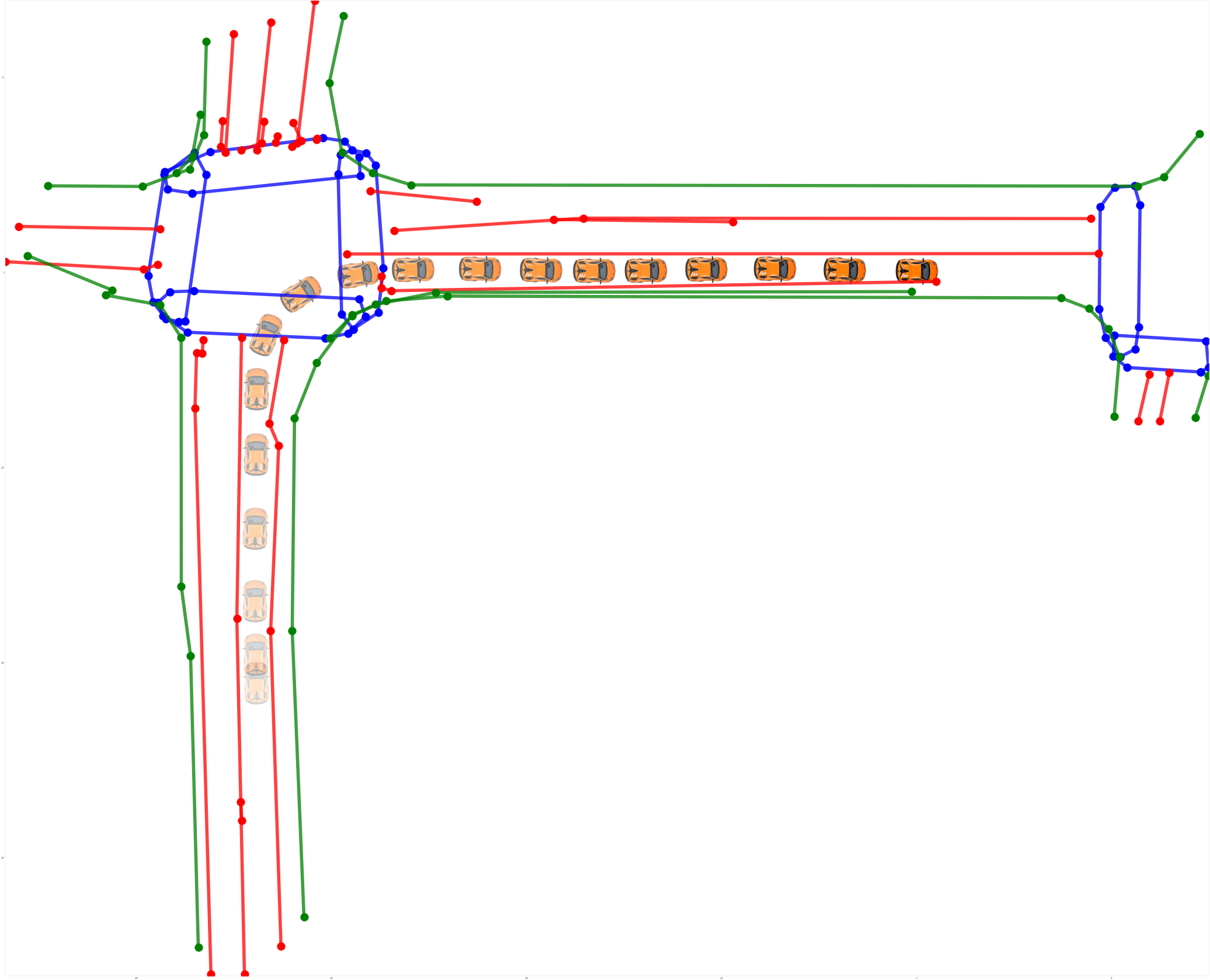}
\end{minipage}
\hfill
\begin{minipage}[t]{0.59\linewidth}
    \centering
    \includegraphics[width=\linewidth]{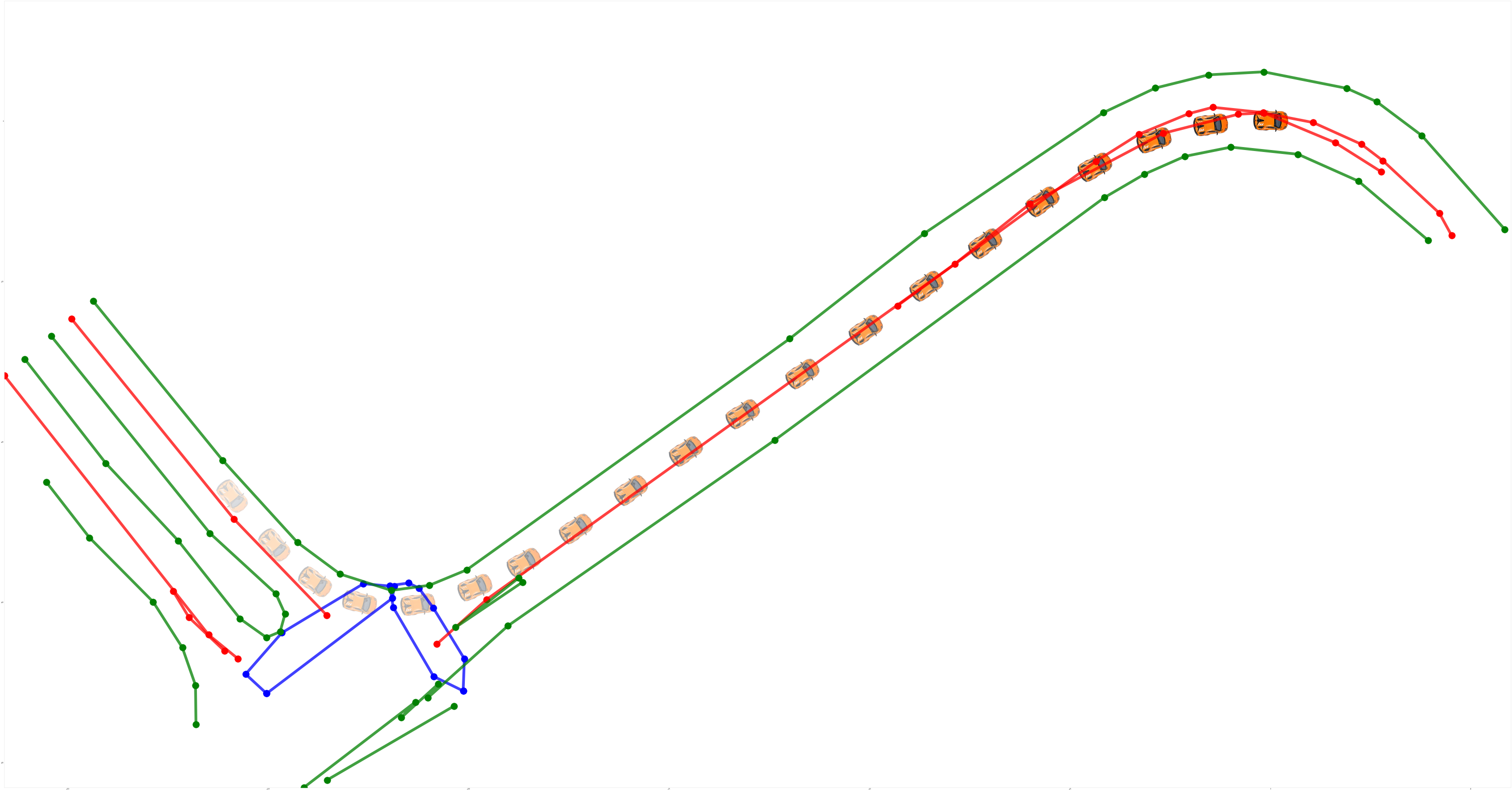}
\end{minipage}
\caption{Qualitative cumulative map reconstructions produced by AerialFusionMapNet for two representative trajectories. Predicted lane boundaries are shown in green, dividers in red, and pedestrian crossings in blue.}
\label{fig:qualitative_examples}
\end{figure*}
\section{AerialFusionMapNet}\label{sec:methodology}

We propose \emph{AerialFusionMapNet}, a fusion-based framework for online \gls{hd} map construction that integrates onboard perception with high-resolution aerial imagery. An overview of the proposed architecture and its two-stage training strategy is illustrated in \Cref{fig:afm_overview}.

We use aerial imagery acquired offline from satellite or aircraft-based surveys and spatially registered to the vehicle-centric coordinate frame as provided by AID4AD~\cite{lengerer2025aid4adaerialimagedata}. 
The aerial patches are metrically consistent with the \gls{bev} representation, ensuring spatial correspondence between overhead and onboard observations.

AerialFusionMapNet consists of four components:
(i) an onboard perception backbone that encodes multi-view inputs into a \gls{bev} feature map,
(ii) an aerial image encoder that extracts representations from overhead imagery,
(iii) a fusion operator that combines aerial and onboard features, and
(iv) a map construction head that predicts vectorized map elements.

\subsection{Fusion Pipeline}

Fusion is performed at the \gls{bev} feature level prior to vectorized map decoding.

Training follows a two-stage procedure that separates aerial representation learning from multimodal fusion.

\textbf{Stage 1: Aerial-only pretraining.}
The aerial encoder is first trained independently using the standard map construction objective without onboard inputs. 
During this stage, scenario-consistent rotation (\gls{scr}) is applied to improve rotational robustness (see \Cref{subsec:aerial_pretraining}). 
This phase enables the encoder to learn a geometrically consistent representation of road structure directly from overhead imagery.

\textbf{Stage 2: Joint aerial–onboard training.}
The full system is subsequently trained with a fixed aerial encoder providing a reference representation. 
Onboard features and the fusion components are optimized under the vectorized map objective while incorporating the \gls{cvs} auxiliary loss to refine the onboard \gls{bev} representation with respect to the aerial reference.

\subsection{Aerial Encoder Architectures}
\label{sec:aerial_encoders}

To study the influence of aerial encoder design, we evaluate four convolutional architectures while keeping the remaining pipeline fixed:

\begin{enumerate}
    \item \textbf{ResUNet}: A ResNet-based encoder~\cite{resnet} with a U-Net-style decoder~\cite{unet}, serving as the baseline used in AID4AD.
    \item \textbf{UNet++}~\cite{zhou2019unetplusplus}: A nested U-Net variant with dense skip connections.
    \item \textbf{ResUNet++}~\cite{resunetpp}: A residual U-Net architecture with enhanced multi-scale aggregation.
    \item \textbf{DeepLabv3+}~\cite{deeplabv3plus}: A high-capacity encoder--decoder architecture employing atrous spatial pyramid pooling.
\end{enumerate}

To ensure compatibility across all evaluated backbones, we extract slightly larger aerial patches at the native spatial resolution such that the input dimensions satisfy encoder-specific architectural constraints. 
The encoder outputs are subsequently center-cropped to the target region of interest
and projected via a stride-4 convolution to match the spatial resolution of the onboard BEV feature map prior to fusion.

\subsection{Aerial-only Pretraining with Scenario-Consistent Rotation}
\label{subsec:aerial_pretraining}

During Stage 1, the aerial encoder is optimized using only overhead imagery and the standard vectorized map objective.

To increase orientation diversity while preserving geometric consistency, we apply scenario-consistent rotation (\gls{scr}) by jointly rotating aerial images and corresponding \gls{bev} annotations around the ego position. The same rotation is applied to all frames within a scene to maintain temporal coherence. This augmentation improves rotational robustness without introducing label misalignment.

The resulting weights initialize the aerial branch for Stage 2 training.

\subsection{Representation Coupling via Cross-View Supervision}
\label{sec:cvs}
In Stage 2, we apply \gls{cvs}~\cite{cvs_bev_surround_teacher} to guide the onboard \gls{bev} representation toward spatially coherent features derived from the perspective-privileged aerial branch.

During training, the aerial encoder provides a reference representation derived from overhead imagery. The onboard features are encouraged to match this reference through a \gls{mse} objective. To account for differences in feature statistics, we introduce a lightweight affine alignment module with learnable per-channel scale and bias parameters applied to the onboard features prior to supervision following~\cite{OneForAll}.

Both representations are L2-normalized, and the \gls{cvs} loss is optimized jointly with the primary map construction objective.

\section{Experimental Setup}

\subsection{Datasets and Aerial Alignment}
All experiments are conducted on the nuScenes dataset~\cite{nuScenes} enriched with precisely registered overhead imagery from AID4AD~\cite{lengerer2025aid4adaerialimagedata}. 
The aerial images are spatially aligned with the vehicle-centric coordinate frame, enabling direct correspondence between overhead and onboard observations. 
For each frame, aerial patches are extracted according to the region of interest used for online map construction.

We follow StreamMapNet~\cite{Streammapnet} and adopt the geographically separated split of Roddick and Cipolla~\cite{Roddick_2020_CVPR}, which reduces geographic overlap between training and validation regions and enables a more realistic evaluation of spatial generalization. This geographically disjoint split serves as the primary benchmark throughout this work. For comparability with prior literature, we additionally report results on the original nuScenes split and provide an overlap-aware analysis to contextualize its performance.

\begin{table*}[t]
\centering
\caption{
Online \gls{hd} map construction performance on the nuScenes geographic split.
We report class-wise AP, \gls{map}, and inference speed where available for two regions of interest.
\textbf{C}: camera images, \textbf{SD}: standard-definition map prior, \textbf{AID}: aerial image data.
* indicates training-only usage.
}
\label{tab:main_results}

\setlength{\tabcolsep}{6pt}

\begin{tabular}{
p{4.0cm}
|
c
|
c
|
S[table-format=2.1]
S[table-format=2.1]
S[table-format=2.1]
S[table-format=2.1]
|
S[table-format=2.1]
S[table-format=2.1]
S[table-format=2.1]
S[table-format=2.1]
}
\toprule

\multicolumn{1}{c}{} &
\multicolumn{1}{c}{} &
\multicolumn{1}{c}{} &
\multicolumn{4}{c}{$60 \times 30$ m} &
\multicolumn{4}{c}{$100 \times 50$ m} \\

\cmidrule(lr){4-7}
\cmidrule(lr){8-11}

\multicolumn{1}{c}{Method} &
\multicolumn{1}{c}{Input} &
\multicolumn{1}{c}{FPS} &
AP$_{\text{ped}}$ & AP$_{\text{div}}$ & AP$_{\text{bound}}$ & \gls{map}
& AP$_{\text{ped}}$ & AP$_{\text{div}}$ & AP$_{\text{bound}}$ & \gls{map} \\

\midrule

StreamMapNet \cite{Streammapnet}
& C & 18.5
& 32.2 & 29.3 & 40.8 & 34.1
& 25.6 & 17.4 & 24.3 & 22.4 \\

MapTracker \cite{maptracker}
& C & 12.5
& 45.9 & 30.0 & 45.1 & 40.3
& 45.9 & 24.3 & 38.4 & 36.2 \\

StreamMapNet-\gls{cvs} \cite{cvs_bev_surround_teacher}
& C + AID\rlap{\textsuperscript{*}} & 18.5
& 41.1 & 29.1 & 45.1 & 38.4
& 40.3 & 25.8 & 30.7 & 32.3 \\

NavMapFusion \cite{monninger2025navmapfusion}
& C + SD & \multicolumn{1}{c|}{--}
& 31.8 & 30.7 & 44.4 & 35.6
& 28.1 & 22.8 & 29.0 & 26.6 \\

AID4AD StreamMapNet \cite{lengerer2025aid4adaerialimagedata}
& C + AID & 18.1
& 59.9 & 32.9 & 53.6 & 48.8
& 73.0 & 27.8 & 41.2 & 47.3 \\

\midrule

\textbf{AerialFusionMapNet (ours)}
& C + AID & 17.7
& \textbf{73.7} & \textbf{35.3} & \textbf{55.2} & \textbf{54.7}
& \textbf{80.9} & \textbf{32.4} & \textbf{45.7} & \textbf{53.0} \\

\bottomrule
\end{tabular}
\end{table*}

\subsection{Region of Interest}

Experiments are conducted on a \(60 \times 30\) meter \gls{roi} centered at the ego vehicle. 
At the native spatial resolution provided by AID4AD (0.15\,m per pixel), this corresponds to aerial patches of \(400 \times 200\) pixels.
To ensure compatibility across encoder architectures, we extract slightly larger patches (224 × 416 in our implementation) while preserving the native metric resolution. 
After feature extraction, encoder outputs are center-cropped to the target \gls{roi} and projected via a stride-4 downsampling module to the \(50 \times 100\) BEV grid used by the onboard Perspective-to-BEV backbone.

To evaluate performance on larger spatial extents, we additionally consider a \(100 \times 50\) meter \gls{roi}. 
For this configuration, aerial inputs are downsampled to 0.2\,m per pixel prior to feature extraction, while the BEV feature grid resolution remains unchanged. 

\subsection{Training Configuration}

Unless stated otherwise, models are trained using the two-stage strategy described in \Cref{sec:methodology}. 

We use AdamW as optimizer with an initial learning rate of \(5\times10^{-4}\). 
Training follows a cosine annealing schedule with linear warmup over the first 500 iterations (warmup ratio \(1/3\)) and a minimum learning rate ratio of \(10^{-3}\). 
Models are trained for 24 epochs with a batch size of 4.

In Stage~2, all trainable components except the aerial encoder are optimized jointly.
When cross-view supervision is enabled, it is incorporated as an auxiliary objective during joint training, without modifying the inference-time architecture.

Following \cite{cvs_bev_surround_teacher}, we use alignment weights of $\lambda_{\text{\gls{cvs}}}=60$ for the \(60 \times 30\) m region of interest and $\lambda_{\text{\gls{cvs}}}=70$ for the \(100 \times 50\) m setting. We additionally evaluate sensitivity to this parameter in our experiments. 

All experiments are conducted under consistent optimization settings to enable a controlled comparison of aerial encoder architectures and training strategies within a shared fusion framework.

\subsection{Baselines and Ablation Protocol}

We evaluate the aerial encoder architectures introduced in \Cref{sec:aerial_encoders} under two training configurations: 
(i) standard end-to-end aerial–onboard fusion, and 
(ii) the proposed two-stage strategy with aerial-only pretraining and cross-view supervision.

For the encoder achieving the highest overall \gls{map} (UNet++, see \Cref{tab:encoder_results}), we conduct additional ablations to isolate the effect of Stage~1 pretraining, representation alignment via \gls{cvs}, and robustness under controlled aerial image offsets.

\subsection{Evaluation Metrics}

Performance is evaluated using \gls{ap} following established practice in \gls{bev} map construction~\cite{maptracker,li2021HDMapNet,liu2022vectormapnet,Streammapnet}. 

For each semantic class, predicted vector elements are matched to ground-truth elements using predefined distance thresholds. 
\gls{ap} is computed independently at each threshold, and the final \acrshort{map} score is obtained by averaging AP across thresholds and semantic classes.
The semantic classes include pedestrian crossings (AP$_{\text{ped}}$), road dividers (AP$_{\text{div}}$) and lane boundaries (AP$_{\text{bound}}$).

For the \(60 \times 30\) m \gls{roi}, we use thresholds of \(\{0.5, 1.0, 1.5\}\,\mathrm{m}\). 
For the \(100 \times 50\) m \gls{roi}, thresholds of \(\{1.0, 1.5, 2.0\}\,\mathrm{m}\) are used.

Inference speed (FPS) is measured on a single NVIDIA A6000 GPU using a batch size of 1.

\section{Results}

\subsection{Main Performance of AerialFusionMapNet}
\label{subsec:mainresults}

We first report the performance of \emph{AerialFusionMapNet} under its best-performing configuration in terms of overall \gls{map}. 
This configuration uses a UNet++ aerial encoder pretrained with \gls{scr} in Stage~1 and frozen during Stage~2, where cross-view supervision is applied with $\lambda_{\text{\gls{cvs}}}=60$.

\Cref{tab:main_results} compares AerialFusionMapNet against prior online \gls{hd} map construction approaches under identical evaluation settings. 
StreamMapNet serves as a camera-only baseline, while MapTracker represents a stronger camera-based model incorporating explicit temporal memory.
NavMapFusion provides an \gls{sd}-map prior baseline on the same geographic split, representing a complementary form of overhead structural guidance.

StreamMapNet-\gls{cvs} improves over the original StreamMapNet by incorporating aerial supervision during training only, while maintaining identical inference-time architecture and speed. 
However, its performance remains below that of MapTracker and the inference-time fusion baselines. 
AID4AD StreamMapNet demonstrates the benefit of aerial-onboard fusion at inference time. 

Building on this paradigm, AerialFusionMapNet achieves the highest overall performance, reaching \textbf{54.7 \gls{map}}.
Qualitative examples are shown in \Cref{fig:qualitative_examples}, illustrating the structural coherence and long-range consistency of the predicted vectorized map elements in complex urban scenes.

The following sections analyze the contribution of individual components of AerialFusionMapNet, including aerial encoder choice, aerial-only pretraining, representation alignment strength, and robustness to aerial image misalignment.

For comparability with prior work, we additionally report results on the original nuScenes split in \Cref{subsec:oldsplit}.

\subsection{Aerial Encoder Study}

\begin{table}
\centering
\caption{Aerial encoder comparison under baseline aerial-onboard fusion and AerialFusionMapNet. Parameter counts refer to the aerial encoder only. FPS is measured for the full inference pipeline.}
\label{tab:encoder_results}
\begin{tabular}{l r c c c}
\toprule
Encoder & \#Params & \gls{map} (Baseline) & \gls{map} (AFM) & FPS \\
\midrule
ResUNet & 7.0M  & 49.8 & 54.3 & \textbf{18.1} \\
DeepLabv3+       & 12.3M & 42.6 & 52.4 & 17.8 \\
UNet++           & 13.0M & 49.7 & \textbf{54.7} & 17.7 \\
ResUNet++        & 24.3M & 41.0 & 50.4 & 16.7 \\
\bottomrule
\end{tabular}
\end{table}

We analyze the influence of aerial encoder architecture under both baseline fusion and the proposed AerialFusionMapNet training strategy. 
Results are summarized in \Cref{tab:encoder_results}, which also reports parameter counts of the aerial encoders.

\paragraph{Baseline aerial–onboard fusion}
Under end-to-end baseline fusion, increasing aerial encoder capacity does not consistently improve performance. 
For example, the 24.3M-parameter ResUNet++ achieves 41.0 \gls{map}, substantially below the 7.0M-parameter ResUNet, which achieves 49.8 \gls{map}. 
Similarly, DeepLabv3+ (12.3M parameters) reaches 42.6 \gls{map}, underperforming smaller architectures. 
These results indicate that encoder capacity alone does not determine fusion quality under naive end-to-end training.

\paragraph{AerialFusionMapNet training strategy}
When trained using the proposed two-stage strategy with frozen aerial encoder and cross-view supervision, performance improves consistently across all architectures. 
ResUNet increases from 49.8 to 54.3 \gls{map}, DeepLabv3+ from 42.6 to 52.4 \gls{map}, UNet++ from 49.7 to \textbf{54.7} \gls{map}, and ResUNet++ from 41.0 to 50.4 \gls{map}.

Notably, larger models do not consistently outperform more compact ones even under the structured training strategy. 
The lightweight 7.0M ResUNet performs on par with higher-capacity alternatives such as the 13.0M UNet++.
UNet++ is used for the main ablations because it achieves the highest overall \gls{map} under our predefined model-selection criterion, while ResUNet provides a nearly equivalent and slightly faster alternative.
This indicates that the structured training design plays a more decisive role than aerial encoder size alone.

\subsection{Impact of SCR-Augmented Aerial Pretraining}

We conduct an ablation study to evaluate how SCR-augmented aerial-only pretraining affects the aerial reference representation and its effectiveness within Stage~2 cross-view supervision.

Using UNet++, we compare two variants of the Stage~2 training pipeline:
(i) initializing the aerial encoder from baseline fusion training, and 
(ii) initializing from aerial-only pretraining with \gls{scr}.

When cross-view supervision is applied using an encoder obtained from baseline fusion training, performance reaches 51.8 \gls{map}. Initializing the aerial encoder via aerial-only pretraining with \gls{scr} further improves performance to 54.7 \gls{map}, corresponding to an additional gain of +2.9 \gls{map}.

These results demonstrate that initializing the aerial encoder via \gls{scr}-augmented aerial-only pretraining noticeably improves the effectiveness of subsequent representation alignment.

\subsection{Sensitivity to Cross-View Supervision Weight}

\begin{figure}[t]
    \centering
    \includegraphics[width=\columnwidth]{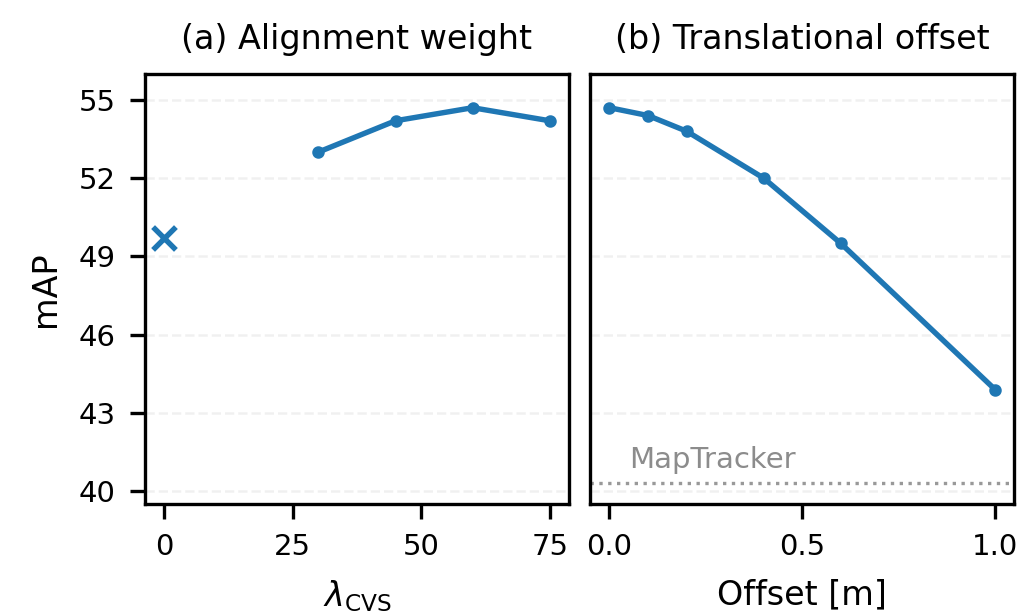}
    \caption{
    Sensitivity analysis of AerialFusionMapNet:
    (a) influence of $\lambda_{\mathrm{CVS}}$;
    (b) performance under translational aerial offsets.
    }
    \label{fig:stability_analysis}
\end{figure}

We analyze the sensitivity of AerialFusionMapNet to the weight of the \gls{cvs} loss. 
Results for the UNet++ configuration are shown in \Cref{fig:stability_analysis}.
Performance remains stable across a broad range of alignment weights between $\lambda_{\text{\gls{cvs}}}=45$ and 75. 
The highest performance of 54.7 \gls{map} is achieved at $\lambda_{\text{\gls{cvs}}}=60$.

\subsection{Robustness to Translational Aerial Misalignment}

We evaluate robustness to synthetic translational misalignment between aerial imagery and the vehicle-centric coordinate frame.
Starting from the AID4AD-aligned aerial imagery, fixed-magnitude translational offsets are applied to the aerial patches prior to fusion. 
For each frame, the offset direction is sampled randomly while the magnitude remains constant. 
Results are shown in \Cref{fig:stability_analysis}.

Performance remains stable for offsets up to 0.2\,m and degrades gradually as the offset increases. 
Competitive performance is maintained up to approximately 0.6\,m.
For reference, AID4AD reports a mean alignment error of approximately 0.16\,m and a maximum error of around 0.6\,m. 
The applied offsets therefore represent additional perturbations beyond the dataset alignment.

\subsection{Results on the Original nuScenes Split}
\label{subsec:oldsplit}

\begin{figure}[t]
\centering
\includegraphics[width=\linewidth]{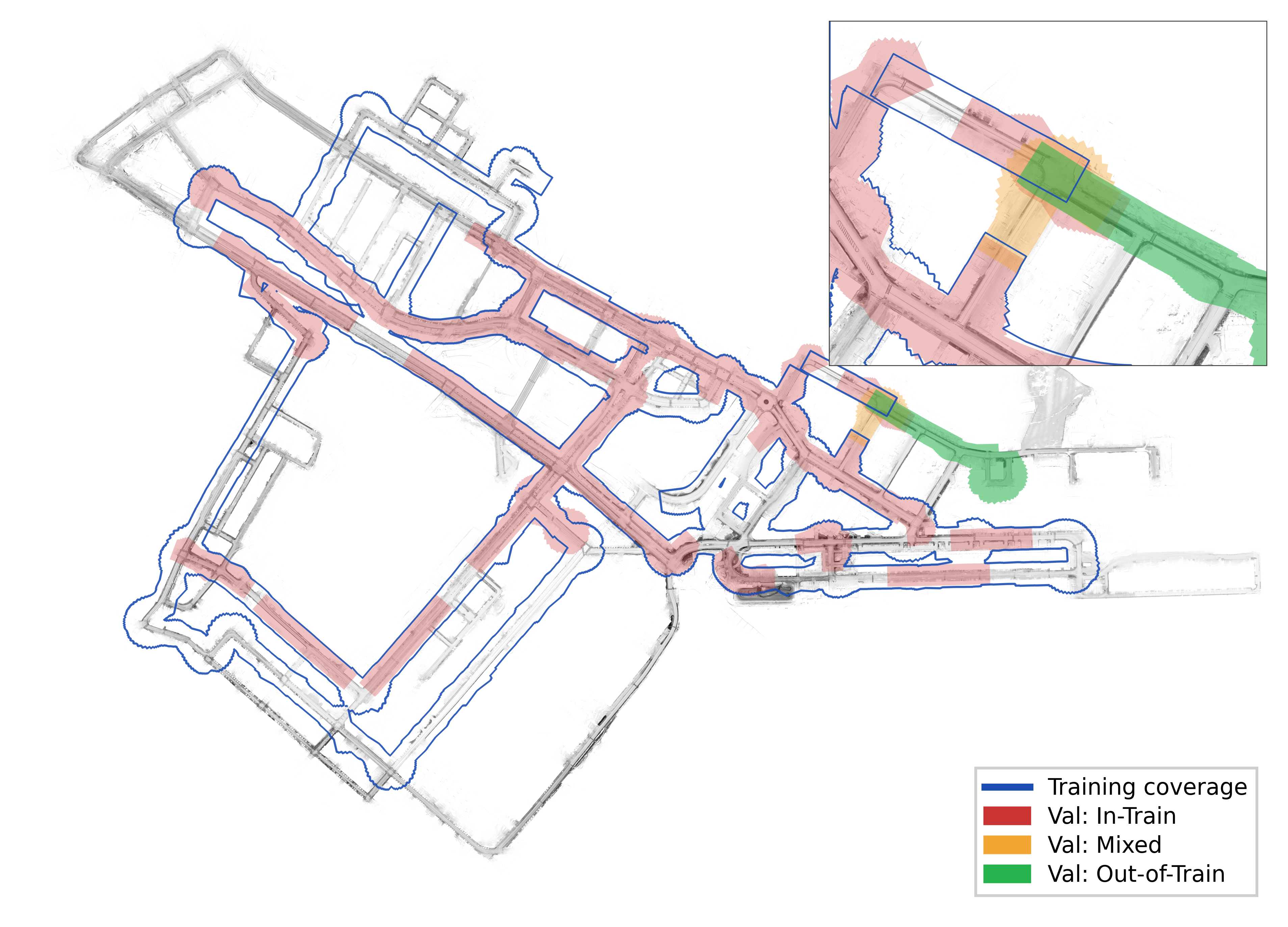}
\caption{
Geographic overlap in the original nuScenes split (Boston-Seaport) using a 100$\times$50 m region of interest.
}
\label{fig:overlap_visualization}
\end{figure}

\begin{table}[t]
\centering
\caption{
Results on the original nuScenes split (60×30 m) for comparability with prior work.
C: camera images, L: LiDAR, AID: aerial image data.
}
\label{tab:oldsplit}
\begin{tabular}{l c c}
\toprule
Method & Input & \gls{map} \\
\midrule
StreamMapNet \cite{Streammapnet} & C & 63.4 \\
MapTracker \cite{maptracker} & C & 71.4 \\
Mask2Map \cite{Mask2Map} & C & 71.6 \\
Mask2Map \cite{Mask2Map} & C + L & 78.4 \\
SatForHDMap \cite{satforhd} & C + AID & 53.7 \\
\midrule
\textbf{AerialFusionMapNet (ours)} & C + AID & \textbf{84.4} \\
\bottomrule
\end{tabular}
\end{table}

\begin{table}[t]
\centering
\caption{Validation performance grouped by geographic overlap with the training split for both regions of interest on the original nuScenes split.}
\label{tab:oldsplit_strat}
\begin{tabular}{l l c c}
\toprule
\gls{roi} & Subset & \#Samples & \gls{map} \\
\midrule
\multirow{4}{*}{60$\times$30 m}
 & In-Train-Area     & 5295 & 87.2 \\
 & Mixed             & 441  & 62.8 \\
 & Out-of-Train-Area & 283  & 54.0 \\
\cmidrule(lr){2-4}
 & Full              & 6019 & 84.4 \\
\midrule
\multirow{4}{*}{100$\times$50 m}
 & In-Train-Area     & 5536 & 90.1 \\
 & Mixed             & 281  & 36.1 \\
 & Out-of-Train-Area & 202  & 39.3 \\
\cmidrule(lr){2-4}
 & Full              & 6019 & 87.5 \\
\bottomrule
\end{tabular}
\end{table}

While the geographically disjoint split serves as our primary benchmark, we additionally report results on the original nuScenes split for comparability with prior work and to assess the effect of geographic overlap (Table~\ref{tab:oldsplit}). AerialFusionMapNet achieves 84.4 \gls{map} for the 60×30 m \gls{roi} and 87.5 \gls{map} for the 100×50 m \gls{roi}, exceeding previously reported approaches under identical evaluation settings, including LiDAR-assisted methods.

The original split exhibits substantial geographic overlap between its training and validation splits \cite{Streammapnet}. Although onboard sensor data is recorded at different timestamps, the aerial image associated with a given spatial location remains identical. For overlapping regions, the aerial branch therefore receives the same static overhead input in both splits, strongly biasing performance toward spatial layouts already present in the training data.

To quantify this effect, we partition the validation data into three overlap-based subsets: In-Train-Area, Mixed, and Out-of-Train-Area. These subsets are defined based on scene-level spatial overlap between each scene’s effective \gls{roi} and the union of training coverage. Overlap is computed at the scene level as the mean \gls{roi} intersection ratio with the training coverage across all frames of a scene. 
Scenes with an average overlap of at least 0.8 are classified as In-Train-Area, scenes with at most 0.2 as Out-of-Train-Area, and the remaining scenes as Mixed.

Figure~\ref{fig:overlap_visualization} visualizes the resulting spatial distribution for the 100$\times$50 m \gls{roi} in the Boston-Seaport location, illustrating the dominance of overlapping regions. The quantitative breakdown is given in Table~\ref{tab:oldsplit_strat}. In both \gls{roi}, the majority of validation samples belong to the In-Train-Area subset (5295/6019 for 60×30 m; 5536/6019 for 100×50 m), confirming that a large fraction of validation samples lie within regions already covered during training.

For the 60×30 m \gls{roi}, Out-of-Train-Area performance (54.0 \gls{map}) is close to the geographically disjoint split (54.7 \gls{map}), indicating meaningful performance outside training coverage at smaller spatial extents. For the 100×50 m \gls{roi}, the gap is larger, with In-Train-Area reaching 90.1 \gls{map} while Mixed and Out-of-Train-Area drop to 36.1 and 39.3 \gls{map}. Despite the limited subset sizes (281 and 202 samples), this consistent gap indicates spatial dependency and should be interpreted as diagnostic evidence for spatial reuse bias rather than as a standalone benchmark for long-range generalization.

Overall, performance on the original split is substantially influenced by validation samples whose spatial support overlaps with training coverage. The performance gap between overlapping and non-overlapping regions is modest for the 60×30 m \gls{roi} but becomes substantially larger for the 100×50 m \gls{roi}.

\section{Discussion}

The results indicate that aerial–onboard fusion performance in online \gls{hd} map construction is closely linked to how aerial and onboard \gls{bev} representations are learned and coupled, rather than to aerial encoder capacity alone. Under standard end-to-end fusion, increasing encoder complexity does not reliably yield improvements, indicating that additional capacity does not necessarily translate into improved fusion performance, either due to representational mismatches during fusion or because the structural information contained in aerial imagery can already be captured effectively by more compact models.

The ablation on \gls{scr}-augmented aerial pretraining further underscores the importance of the aerial reference representation for effective multimodal coupling. Robustness experiments show that the learned aerial–onboard interaction remains stable under moderate translational offsets beyond the reported dataset alignment error while still benefiting from precise registration. This indicates that the fusion mechanism exploits spatial structure in a locally consistent manner rather than relying on pixel-perfect correspondence.

Taken together, these findings highlight representation quality and cross-view compatibility as central factors in aerial–onboard map construction, while confirming that increasing aerial encoder capacity alone is insufficient to guarantee improved performance.

The overlap-aware evaluation on the original split further shows that performance can be strongly influenced by spatial reuse when training and validation regions overlap. In aerial–onboard fusion settings, where identical overhead imagery may be reused across splits, such effects can substantially shape reported performance.

From an application perspective, the results suggest that carefully structured training strategies can mitigate the need for high-capacity aerial encoders while preserving strong mapping accuracy. This is particularly relevant for deployment-constrained automated driving systems, where computational efficiency and model compactness remain critical design considerations.

\section{Conclusion}

This work presented AerialFusionMapNet, a fusion-based framework for online HD map construction that integrates overhead aerial imagery with onboard perception in a unified \gls{bev} representation.

Our experiments demonstrate that aerial–onboard fusion performance is governed less by aerial encoder capacity than by how aerial representations are learned and coupled with onboard features. While naive end-to-end fusion does not consistently benefit from increased encoder complexity, the proposed two-stage strategy combining aerial-only pretraining with \acrlong{scr} and representation-level alignment via \acrlong{cvs} yields stable and architecture-agnostic performance gains.

Beyond overall accuracy improvements, our analysis reveals that geographic overlap between training and validation data can substantially influence evaluation outcomes when aerial imagery is fused at inference time. This underscores the importance of geographically separated benchmarks for assessing spatial generalization in multimodal map construction.

Furthermore, the proposed training design exhibits robustness to moderate translational misalignment of aerial imagery, indicating that effective cross-view fusion remains feasible under realistic registration uncertainty.

Overall, the findings emphasize representation quality and cross-view compatibility as central factors in aerial–onboard map construction and provide a principled foundation for future research on multimodal perception and mapping in automated driving.

\section*{Acknowledgment}
This work is supported by the NeMo.bil project 19S23003, which
is funded by the Federal Ministry for Economic Affairs and Energy of Germany.
The authors gratefully acknowledge the scientific support and HPC resources provided by the Erlangen National High Performance Computing Center (NHR@FAU) of the Friedrich-Alexander-Universität Erlangen-Nürnberg (FAU) under the BayernKI project v153eb. BayernKI funding is provided by Bavarian state authorities.

\bibliographystyle{IEEEtran}
\bibliography{main}

\end{document}